\documentclass[11pt,a4paper]{article}
\usepackage[hyperref]{eacl2021}
\usepackage{times}
\usepackage{latexsym}
\usepackage{graphicx}

\usepackage{microtype}

\usepackage{textcomp}

\aclfinalcopy % Uncomment this line for the final submission
\def\aclpaperid{297} %  Enter the acl Paper ID here

\newcommand{\statsign}{\textsuperscript{$\dagger$}}
\newcommand{\statsignbis}{\textsuperscript{*}}

\title{CTC-based Compression for Direct Speech Translation}

\author{Marco Gaido\textsuperscript{$\dagger$, $*$}, Mauro Cettolo\textsuperscript{$\dagger$}, Matteo Negri\textsuperscript{$\dagger$}, Marco Turchi\textsuperscript{$\dagger$} \\
  \textsuperscript{$\dagger$}Fondazione Bruno Kessler \\
  \textsuperscript{$*$}University of Trento \\
  \texttt{\{mgaido,cettolo,negri,turchi\}@fbk.eu} \\}

\date{}

\begin{document}
\maketitle
\begin{abstract}
Previous studies demonstrated that a dynamic phone-informed compression of
the input audio is beneficial for speech translation (ST).
However,
they required a dedicated model for phone recognition
and did not test this solution for direct ST,
in which a single model translates the input audio into the target
language without intermediate representations.
In this work, we propose 
the first method able
to perform a dynamic compression
of the input in direct ST models. In particular,
we exploit the Connectionist Temporal Classification (CTC)
to compress the input sequence according to its phonetic
characteristics.
Our experiments demonstrate that 
our
solution brings a 1.3-1.5 BLEU improvement over a strong baseline
on two language pairs (English-Italian and English-German), contextually reducing
the memory footprint by more than 10\%.
\end{abstract}

\section{Introduction}

Speech translation (ST) is the process that converts utterances in one language into text in another language.
Traditional approaches to ST consist of 
separate
modules, each dedicated to an easier sub-task,
which are eventually integrated in a so-called \textit{cascade} architecture \cite{StentifordSteer88,Waibel1991b}.
Usually, 
its main components are an automatic speech recognition (ASR) model - which generates the transcripts from the audio - and a machine
translation (MT) model - which translates the transcripts into the target language. 
A newer approach is 
\textit{direct} ST, in which
a single model performs the whole task without intermediate 
representations \cite{berard_2016,weiss2017sequence}.
The main advantages of direct ST systems are:
\textit{i)} the access to information
not present in the text
(e.g. prosody, vocal characteristics of the speaker) during the translation phase, 
\textit{ii)} a reduced latency, 
\textit{iii)}  a simpler and easier to manage architecture (only one model has to be maintained), which \textit{iv)} avoids error propagation across components.

In both paradigms (cascade and direct),
the audio is commonly represented as a sequence of
vectors obtained with a Mel
filter bank.
These vectors are collected with a high frequency, typically one every 10 ms.
The resulting sequences are much longer than
the corresponding textual ones (usually by a factor of \texttildelow 10).
The sequence length is problematic both for RNN \cite{Elman90findingstructure}
and Transformer \cite{transformer} architectures.
Indeed, RNNs fail to represent long-range dependencies \cite{bengio-1993-RNN-long-term} and
the Transformer has a quadratic memory complexity
in the input sequence length,
which makes training on long sequences prohibitive due to its memory footprint.
For this reason, architectures proposed for direct 
ST/ASR
reduce the input length either with convolutional layers \cite{berard_2018,di-gangi-etal-2019-enhancing} or by stacking and downsampling
consecutive samples
\cite{sak2015fast}.
However, these fixed-length reductions of the input sequence assume that
samples carry the same amount of information.
This does not necessarily hold true, as
phonetic features vary at a different speed in time and frequency in the audio signals.

Consequently, researchers have studied how to reduce the input length
according to dynamic criteria based on the audio content.
\citet{salesky-etal-2019-exploring-ph} 
demonstrated that a phoneme-based compression of the input frames
yields significant gains compared to fixed
length reduction.
Phone-based and linguistically-informed compression also proved to be useful in the context of visually grounded speech \cite{havard2020catplayinginthesnow}.
However, \citet{salesky-black-2020-phone} questioned the 
approach, claiming that the addition of phone features without
segmentation and compression of the input is more effective.

None of these works is a direct ST solution, as they all require a separate
model for phone recognition and intermediate representations.
So, they: \textit{i)} are affected by \textit{error propagation} (\citealt{salesky-black-2020-phone} show in fact that lower quality in phone recognition significantly degrades final ST performance), \textit{ii)} have higher latency and \textit{iii)} a more complex architecture.
A direct model with phone-based multi-task training was introduced by \citet{Jia2019} for speech-to-speech translation, but they neither  compared with a training using transcripts nor investigated dynamic compression.

In this paper, we 
explore
the usage of phones and
dynamic content-based input compression for direct ST 
(and ASR). Our goal is an input reduction 
that, limiting the amount of redundant/useless information, yields better performance and lower memory consumption at the same time.
To this aim, we propose to exploit the
Connectionist Temporal Classification (CTC) \cite{Graves2006ConnectionistTC}
to add phones prediction in a multi-task training and compress the sequence accordingly.
To disentangle the contribution of the introduction of phone recognition and the compression based on it, we compare 
against similar trainings leveraging transcripts instead of phones.
Our results show that phone-based multi-task training with sequence
compression improves over a strong
baseline by up to 1.5 BLEU points
on two language pairs 
(English-German and English-Italian), with a memory footprint reduction of at least 10\%.

\section{CTC-based Sequence Compression}

The CTC algorithm is usually employed for training a model to predict an output sequence of variable length that is shorter than the input one. This 
is the case of 
speech/phone recognition, as the input is a long sequence of audio samples, while the output is the sequence of uttered symbols (e.g. phones, sub-words), which is significantly shorter.
In particular, for each time step, the CTC produces a probability distribution over the possible target labels augmented with a dedicated \texttt{<blank>} symbol representing the absence of a target value.
These distributions are then 
exploited to compute the probabilities of different sequences, in which consecutive equal predictions are collapsed and \texttt{<blank>} symbols are removed. Finally, the resulting sequences are compared with the target sequence.

\begin{figure}[htbp]
\centering
\includegraphics[width=7cm]{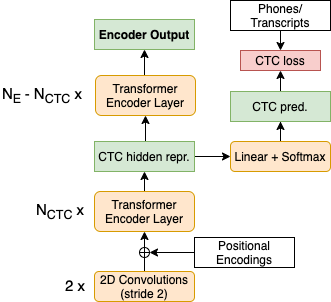}

\caption{Encoder architecture with CTC loss.}
\label{fig:ctc_nored}
\end{figure}

Adding an auxiliary CTC loss to the training of
direct ST and acoustic ASR models has been shown
to improve performance \cite{kim-et-al-2017-joint,bahar2019comparative}.
In these works, the CTC
loss is computed against the
transcripts 
on the encoder output to  
favour model convergence.
Generally,
the CTC loss can be added to the output of any encoder  
layer, as in Figure \ref{fig:ctc_nored} where  the hyper-parameter $N\textsubscript{CTC}$ indicates the number of the layer at which the CTC is computed.
Formally, the final loss function is:

\begin{equation}
    \lambda = CTC(E\textsubscript{N\textsubscript{CTC}}) + CE(D\textsubscript{N\textsubscript{D}})
\end{equation}

\noindent
where $E\textsubscript{x}$ is the output of the $x$-th encoder layer, $D\textsubscript{N\textsubscript{D}}$ is the decoder output, $CTC$ is the CTC function,
and $CE$ is the label smoothed cross entropy.
If $N\textsubscript{CTC}$ is equal to the number of encoder layers ($N\textsubscript{E}$),
the CTC input is
the encoder output.
We consider this solution as our baseline and we also test it with phones as target.

As shown in Figure \ref{fig:ctc_nored}, we use as model a Transformer, whose encoder layers
are preceded by two 2D convolutional layers that reduce the input size by a factor of 4.
Therefore, the CTC produces a prediction every 4 input time frames.
The sequence length reduction is necessary both because it makes possible the training (otherwise
out of memory errors would 
occur) and to have a fair comparison with modern state-of-the-art models.
A logarithmic distance penalty \cite{di-gangi-etal-2019-enhancing}
is added to all the Transformer encoder layers.

\begin{figure}[htbp]
\centering
\includegraphics[width=7cm]{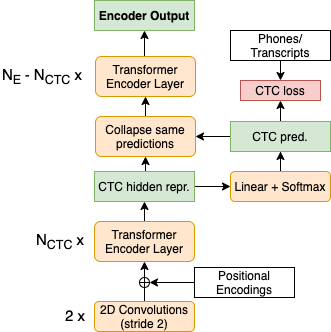}
\caption{Encoder architecture with CTC compression.}

\label{fig:ctc_red}
\end{figure}

Our proposed architecture is represented in Figure \ref{fig:ctc_red}.
The difference with the baseline is the introduction of an additional block (\textit{Collapse same predictions}) that exploits
the CTC predictions to compress the input elements (vectors). Hence, in this case the CTC
does not only help model convergence, but it also defines variable-length segments representing the same content. So, dense audio portions can be given more importance, while redundant/uninformative vectors can be compressed. This allows the following encoder layers and the decoder to attend to useful information without being ``distracted'' by noisy elements.
The architecture is a direct ST solution
as there is a single model whose
parameters are optimized together without intermediate representations.
At inference time, the only input
is the audio
and the model
produces the translation into the target language
(contextually generating the transcripts/phones with the CTC).

We compare three techniques 
to compress the consecutive vectors
with the same CTC 
prediction:

\begin{itemize}
    \item \textbf{Average.} The vectors to be collapsed together are averaged. As there is only a linear layer between the CTC inputs and its predictions, the vectors in each group are likely to be similar, so the compression should not remove much information.
    
    \item \textbf{Weighted.} The vectors are averaged but the weight of each vector depends on the confidence (i.e. the predicted probability) of the CTC prediction. This solution is meant to  give less importance to vectors whose phone/transcript is not certain.
    
    \item \textbf{Softmax.} In this case, the weight of each vector is obtained by computing the \texttt{softmax} of the CTC predicted probabilities. The idea is to propagate information (nearly) only through a single input vector (the more confident one) for each group.
\end{itemize}

\section{Data}

We experiment with
MuST-C \cite{mustc},
a multilingual ST corpus built from 
TED talks.
We focus on 
the English-Italian (465 hours) and English-German (408 hours) sections.
For each set (train, validation, test),
it contains the audio files, the
transcripts,
the
translations and a YAML file
with 
the  start time and duration of the segments.

In addition, we extract the phones using Gentle.\footnote{\url{https://lowerquality.com/gentle/}}
Besides aligning the transcripts with the audio, Gentle returns the start and end time for each recognized word,
together with 
the corresponding phones.
For the words not recognized in the audio, Gentle does not provide the phones, so we lookup their
phonetic transcription
on the VoxForge\footnote{\url{http://www.voxforge.org/home}} dictionary.
For each sample in the corpus,
we rely on the YAML file and the alignments generated by Gentle 
to get all the words (and phones) belonging to it.
The phones have a suffix indicating the position in a word 
(at the end, at the beginning, in the middle or standalone).
We also generated a version without the suffix (we refer to it as
\texttt{PH W/O POS} in the rest of the paper).
The resulting dictionaries contain respectively 144 and 48 symbols.

\section{Experimental Settings}

Our Transformer layers have 8 attention heads,
512 features for the attention and 2,048 hidden units in FFN.
We set a 0.2 dropout and include \textit{SpecAugment} \cite{Park_2019} in
our trainings. We optimize label smoothed cross entropy \cite{szegedy2016rethinking}
with 0.1 smoothing factor using Adam \cite{adam} (betas \textit{(0.9, 0.98)}).
The learning rate increases linearly from 3e-4 to 5e-3 for 4,000 updates,
then decays with the inverse square root.
As we train on 8 GPUs with mini-batches of 8 sentences
and we update
the model every 8 steps,
the resulting batch size is 512.
The audio is pre-processed performing speaker normalization and extracting 40-channel Mel filter-bank features per frame.
The text is tokenized into subwords with 1,000 BPE merge rules \cite{sennrich2015neural}.

As having more encoder layers than decoder layers has been shown to be beneficial \cite{potapczyk-przybysz-2020-srpols,gaido-etal-2020-end}, we use 8 Transformer encoder layers and 6 decoder layers for ASR and 11 encoder and 4 decoder layers for ST unless stated otherwise.
We train until the model does not improve on the validation set for 5 epochs
and we average the last 5 checkpoints.
Trainings were performed on K80 GPUs and lasted \texttildelow 48 hours (\texttildelow 50 minutes per epoch). Our implementation\footnote{Available at \url{https://github.com/mgaido91/FBK-fairseq-ST/tree/eacl2021}.} is based on Fairseq \cite{ott-etal-2019-fairseq}.

We evaluate performance with WER for ASR and
with BLEU \cite{papineni-2002-bleu}\footnote{To be comparable with previous works.} and SacreBLEU \cite{post-2018-call}\footnote{The version signature is: \texttt{BLEU+c.mixed+\#.1+s.exp+tok.13a+v.1.4.3}.} for ST.

\begin{table}[h]
\small
\centering
\begin{tabular}{lcc}
\hline
& \textbf{WER ($\downarrow$)} & \textbf{RAM (MB)} \\
\hline
Baseline - 8L EN  & 16.0 & 6929 (1.00) \\
8L PH & \textbf{15.6} & 6661 (0.96) \\
\hline
2L PH AVG & 21.2 & 3375 (0.49) \\
4L PH AVG & 17.5 & 4542 (0.66) \\
8L PH AVG & 16.3 & 6286 (0.91) \\ 
8L PH W/O POS. AVG &  16.4 & 6565 (0.95) \\
8L EN AVG & 16.3 & 6068 (0.88) \\
\hline
\end{tabular}
\caption{\label{tab:asr-results}
Results on ASR using the CTC loss with transcripts and phones as target.
\texttt{AVG} indicates that sequence is compressed averaging the vectors.}
\end{table}

\begin{table*}[t]
\small
\centering
\begin{tabular}{lccc|ccc}
\hline
 & \multicolumn{3}{c}{\textbf{en-it}} & \multicolumn{3}{c}{\textbf{en-de}} \\
 & \textbf{BLEU ($\uparrow$)} & \textbf{SacreBLEU ($\uparrow$)} & \textbf{RAM (MB)} & \textbf{BLEU ($\uparrow$)} & \textbf{SacreBLEU ($\uparrow$)} & \textbf{RAM (MB)} \\
\hline
\cite{di-gangi-etal-2020-target} & 20.1 & - & - & 19.1 & - & - \\
\hline
Baseline - 8L EN & 22.1 & 21.8 & 9624 (1.00) & 20.4 & 20.5 & 9166 (1.00) \\
8L PH & 22.6\statsignbis & 22.3\statsignbis & 9567 (0.99) & 21.6\statsignbis & 21.6\statsignbis & 9190 (1.00) \\
\hline
2L PH AVG & 20.2 & 20.0 & 5804 (0.60) & 17.8 & 17.8 & 4484 (0.49) \\
4L PH AVG & 21.6 & 21.3 & 6193 (0.64) & 20.1 & 20.2 & 5186 (0.57) \\
8L PH AVG & \textit{23.2\statsign} & \textit{22.8\statsign} & 8554 (0.89) &  \textit{21.8\statsignbis} & \textit{21.9\statsignbis} & 7348 (0.80) \\
\hline
8L PH WEIGHTED & 22.7\statsignbis & 22.5\statsignbis & 7636 (0.79) & 21.7\statsignbis & 21.8\statsignbis & 7380 (0.81) \\
8L PH SOFTMAX & 22.6\statsignbis & 22.3\statsignbis & 7892 (0.82) & \textit{21.8\statsignbis} & \textit{21.9\statsignbis} & 7436 (0.81) \\
\hline
8L PH W/O POS. AVG & 22.2 & 22.0 & 7451 (0.77) & 21.5\statsignbis & 21.6\statsignbis & 7274 (0.79) \\
8L EN AVG & 22.2 & 21.9 & 8287 (0.86) & 20.6 & 20.7 & 7143 (0.78) \\
\hline
8L PH AVG (14+6L) & \textbf{23.4\statsign} & \textbf{23.2\statsign} & 8658 (0.90) & \textbf{21.9\statsign} & \textbf{22.0\statsign} & 7719 (0.84) \\
\hline
\end{tabular}
\caption{\label{tab:st-results}
Results using the CTC loss with transcripts and phones as target.
\texttt{AVG}, \texttt{WEIGHTED} and \texttt{SOFTMAX} indicate the compression method.
If none is specified, no compression is performed.
The symbol ``\statsignbis'' indicates improvements that are
statistically significant with respect to the baseline.
``\statsign'' indicates
statistically significant gains with respect to \texttt{8L PH}.
Statistical significance is computed according to \cite{koehn-2004-stat} with $\alpha=0.05$.
Scores in \textit{italic} indicate the best models among those with equal number of layers.}
\end{table*}

\section{Results}

\subsection{ASR}

We first tested whether ASR benefits from the usage of phones and sequence compression.
Table \ref{tab:asr-results} shows that having phones instead of English transcripts (Baseline - 8L EN) as target of the CTC loss (\texttt{8L PH}) without compression
is beneficial.
When compressing the sequence, there is little difference according to the target used
(\texttt{8L PH AVG}, \texttt{8L PH W/O POS. AVG}, \texttt{8L EN AVG}).
However, the compression causes a 0.3-0.5 WER performance degradation
and a 12-5\% saving of RAM.
Moving the compression
to previous layers (\texttt{4L PH AVG}, \texttt{2L PH AVG})
further decreases the output quality 
and
the RAM usage.
We can conclude that compressing the input sequence harms ASR performance,
but might be useful if RAM usage is critical and should be traded off with performance.

\subsection{Direct ST}

In early experiments, we pre-trained
the first 8 layers of the ST encoder with that of the ASR model, adding three \textit{adapter} layers \cite{bahar2019comparative}. We realized that ASR pre-training was not useful (probably because ASR and ST data are the same), so we report results without pre-training.

As we want to ensure that our results are not biased by a poor baseline, we compare with \cite{di-gangi-etal-2020-target},
which uses the same framework and similar settings.\footnote{We 
acknowledge
that better results 
have been 
published  in a contemporaneous 
paper
by \newcite{inaguma-etal-2020-espnet}. Besides the contemporaneity issue,
our results are not comparable with theirs, 
as they use:
\textit{i)} a different architecture built on ESPnet-ST (a newer
framework that, alone, outperforms Fairseq), \textit{ii)} 
higher dimensional input features (83 vs 40 dimensions),
\textit{iii)} data augmentation,
and
\textit{iv)} pre-training techniques.}
As shown in Table \ref{tab:st-results}, our strong baseline (\texttt{8L EN}) outperforms \cite{di-gangi-etal-2020-target} by 2 BLEU on en-it and 1.3 BLEU on en-de.

As in ASR, replacing the transcripts with phones as target for the CTC loss (\texttt{8L PH})
further improves respectively by 0.5 and 1.2 BLEU.
We first explore the introduction of the compression at different layers. Adding it to the 8\textsuperscript{th} layer (\texttt{8L PH AVG}) enhances the translation quality by 0.6 (en-it) and 0.2 (en-de) BLEU, with the improvement on en-it being statistically significant over the version without CTC compression. Moving it to
previous layers (\texttt{4L PH AVG}, \texttt{2L PH AVG}) causes performance
drops, suggesting that many layers are needed to extract useful phonetic information.

Then, we compare the different compression policies: \texttt{AVG} outperforms (or matches) \texttt{WEIGHTED} and \texttt{SOFTMAX} on both languages.
Indeed, the small weight these two methods assign to some vectors likely
causes an information loss and prevents a proper gradient propagation for the corresponding input elements.

Finally, we experiment with different CTC targets, but both
the phones without
the position suffix
(\texttt{8L PH W/O POS. AVG}) and the transcripts (\texttt{8L EN AVG})
lead to lower scores.

The different results between ASR and ST can be explained by the nature of the two tasks: extracting content knowledge is critical for ST but not for ASR, in which a compression can hide details that are not relevant
to extrapolate meaning,
but needed
to generate precise transcripts.
The RAM savings are higher than in ASR as there are 3 more
layers.
On the 8\textsuperscript{th}
layer, they range from 11\% to 23\% for en-it, 16\% to 22\% for en-de.
By moving the
compression
to previous layers, we can trade performance for RAM requirements, saving up to 50\% of the memory.

We also tested whether we can use the saved RAM to add more layers
and improve the translation quality.
We added 3 encoder
and 2 decoder layers: this (\texttt{8L PH AVG (14+6L)}) results in small
gains (0.2 on en-it and 0.1 on en-de), but the additional memory required is also small
(the RAM usage is still 10-16\% lower than the baseline).
The improvements are statistically significant with respect to the
models without compression (\texttt{8L PH}) on both language pairs.
When training on more data, the benefit of having deeper networks might be higher, though, and this solution allows to increase the number of layers without a prohibitive memory footprint. We leave this investigation 
for future works,
as experiments on 
larger training corpora
are out of the scope of this paper.

\section{Conclusions}

As researchers' focus is shifting from cascade to direct solutions due to the advantages of the latter,
we proposed
a technique of dynamic sequence-length reduction for direct ST.
We showed that averaging the vectors corresponding to the same phone prediction according to the CTC
improves the translation quality
and reduces 
the
memory
footprint, allowing for training deeper models.
Our best model outperforms a strong baseline,
which 
uses transcripts in a multi-task training,
by 1.3 (en-it) and 1.5 (en-de) BLEU, reducing memory usage by 10-16\%.

\section*{Acknowledgments}

This work is part of the ``End-to-end Spoken Language Translation in Rich Data Conditions'' project,\footnote{\url{https://ict.fbk.eu/units-hlt-mt-e2eslt/}} which is financially supported by an Amazon AWS ML Grant. The authors also wish to thank Mattia Antonino Di Gangi for the insightful discussions on this work.

\bibliography{eacl2021}
\bibliographystyle{acl_natbib}

\end{document}